# An Improved Person Re-identification Method by light-weight convolutional neural network

Sajad Amouei Sheshkal
M.Sc. Graduate,
Deep Learning Research Lab,
Department of Computer Engineering,
Faculty of Engineering,
College of Farabi,
University of Tehran, Iran
sajad.amouei@ut.ac.ir

Kazim Fouladi-Ghaleh
Assistant Professor,
Deep Learning Research Lab,
Department of Computer Engineering,
Faculty of Engineering,
College of Farabi,
University of Tehran, Iran
kfouladi@ut.ac.ir

Hossein Aghababa
Assistant Professor, Quantum
Computation & Communication Lab,
Department of Computer Engineering,
Faculty of Engineering,
College of Farabi,
University of Tehran, Iran
aghababa@ut.ac.ir

*Abstract*—Person Re-identification is defined as a recognizing process where the person is observed by non-overlapping cameras at different places. In the last decade, the rise in the applications and importance of Person Re-identification for surveillance systems popularized this subject in different areas of computer vision. Person Re-identification is faced with challenges such as low resolution, varying poses, illumination, background clutter, and occlusion, which could affect the result of recognizing process. The present paper aims to improve Person Re-identification using transfer learning and application of verification loss function within the framework of Siamese network. The Siamese network receives image pairs as inputs and extract their features via a pre-trained model. EfficientNet was employed to obtain discriminative features and reduce the demands for data. The advantages of verification loss were used in the network learning. Experiments showed that the proposed model performs better than state-of-the-art methods on the CUHK01 dataset. For example, rank5 accuracies are 95.2% (+5.7) for the CUHK01 datasets. It also achieved an acceptable percentage in Rank 1. Because of the small size of the pre-trained model parameters, learning speeds up and there will be a need for less hardware and data.

## I. Introduction

The development of surveillance equipment and the greater demands for public safety caused more IP cameras to be installed in public places, such as parks, airports, university campuses, etc. [1] Person Re-identification is usually examined as an image retrieval problem, with the aim of matching the images of the same pedestrian captured by different cameras [2]. The image of the pedestrian in question is entered as a query, and Person Re-identification reveals that which cameras have also captured the same pedestrian [1].

There has been considerable progress recently in the Person Re-identification studies. The reasons for such progress include the larger datasets and training the convolutional neural networks (CNN) in pedestrian descriptors. Despite the achievements by the computer vision researchers, there are still unresolved challenges [3]. Fig. 1 exemplifies a Person Re-identification challenge.

Person Re-identification has limited labeled images. Existing methods usually employ pre-trained models on larger datasets, and then adjust the model to the target dataset; this is called transfer learning that may deliver unfavorable results [2]. Verification models receive the image pairs and give them similarity points [4]. If the person in the image pair is the same, the images will be close together in the feature space; otherwise, they will be far apart. A disadvantage of verification models is that they do not used image labels; they compare each image pair, ignoring their relationship with other images [1].

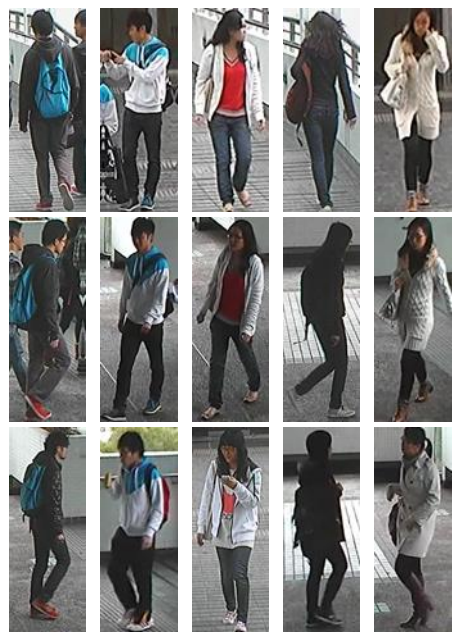

Fig. 1. An example of an inter-class challenge related to the CUHK01 data set. First row: query images. Second row: Correctly recognized images. Last row: Incorrectly recognized images.

This paper intends to create an end-to-end network that gives one similarity point to both of the pedestrian images. The proposed method uses EfficientNet network as the feature extractor to find the inter-class and intra-class relationships of deep features in the higher layer. Finally, the extracted features are monitored by the verification loss function.

The good performance of deep learning caused the researchers to use it to determine the apparent features and distance metrics for Person Re-identification [5]. Deep Person Re-identification aims to use the emergent CNNs and automatically extract the features of image pairs; this way the likelihood of the matching in both images grows. Therefore, features representation and distance metric combine to train an integrated end-to-end system. This combination turns the convolutional optimization into a combined process for training [6].

Since Siamese networks started to be used for Person Re-identification, the proposed models have focused on the development of feature extraction and learning distance metric [1]. Paper [7] used feature extraction and LSTM network based on the Siamese architecture. In metric learning methods, Paper [8] presented a multi-task deep neural network (DNN) that performs ordering and classification jointly; it developed a multiple-task architecture to establish supremacy over other methods. Paper [9] improved the metric learning distinction using the quadratic loss function and hard samples and enhanced the Person Re-identification generalizability.

Paper [10] employed the two-stream Siamese network in order to identify pedestrians in images with low resolution. The part-matching layer distinguishes the FPNN architecture [2] from the previous one, which multiplies convolutional responses by different horizontal stripes. The FPNN model used ImprovedReID [11] method to determine the differing features of the nearest neighbors and improve the input, comparing the features of the input image with local features of its neighbors. Person Re-identification can be considered a type of identification or verification. Some studies, [4], [8] and [12]–[16] have used identification and verification in their proposed methods.

This paper is organized in four sections: Section 2 presents the proposed method. Section 3 presents the results and discussion, conclusion and some suggestions for future works appear in the fourth section.

## II. PROPOSED METHOD

The proposed network is based on a Siamese convolutional model. In the beginning of the network, a pre-trained model, i.e. EfficientNet, extracts the features from input images. After extracting high-level features, they enter the Square Layer to be compared. At the end of the network, Verification Loss is used to determining the similarity. Fig. 2 shows the proposed network architecture.

The objective of the proposed model was to learn the representation of the local features of the input image pairs and determine the similarity points. This way the network could learn discriminative features in order to classify input images from different classes. First, the images enter the network for feature extraction. There is a multi/N-modal feature descriptor after the last layer of the pre-trained network. Afterwards, the extracted features of the higher layer enter a parameterless square [4] for comparison. This layer takes the matrix as an input and delivers another matrix as an output after subtracting and squaring. There is a drop-out layer in the network to avoid the overfitting. At the end of the network is the verification loss, which determines how many percent of the two input images belong to one person. We have provided our code publicly available[1].

### A. EfficientNet

Google researchers trained EfficientNet model on the ImageNet dataset that contains 2000 classes and over 14 million images. EfficientNet can be subsumed under CNNs. EfficientNet that comes in 8 versions, from *B*0 (lowest) to *B*7 (highest) differing in number of parameters and accuracy, is much more efficient than the previous models. The best version of EfficientNet is 8.4 times smaller and 6.1 times quicker than the best pre-trained models [17].

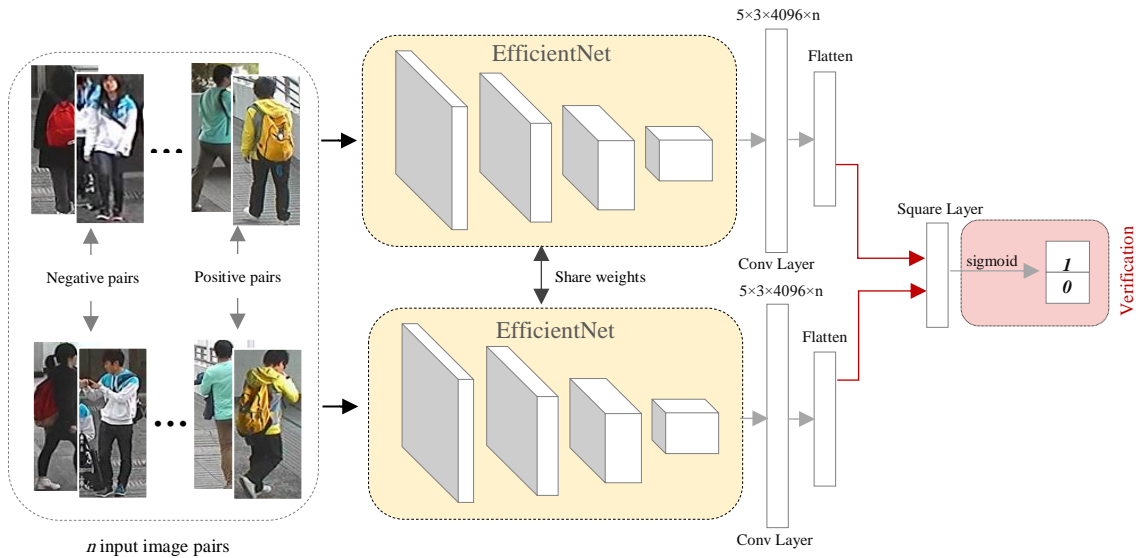

Fig. 2. Architecture of the proposed method

---

[1] https://github.com/sajadamouei/Person-Re-ID-with-light-weight-network

The main idea for the development of EfficientNet was to make changes in three dimensions of depth, width, and resolution at a fixed rate. In previous pre-trained models, one of the said dimensions rose above others. Fig. 3 displays the difference between a change on one dimension and a change on several dimensions at a fixed rate. Changes in only one of the dimensions hinder the performance of networks; this shows the limitation of unidimensional scaling.

MBConv constitutes the major building block of EfficientNet, which is an inverted conv bottleneck, its first version being used in MobileNetV2 model. They establish a shortcut link between the beginning and end of a convolutional block. Convolutional depth and point first expand the feature maps and then contract them. Narrow layers are connected via shortcut links, while the broad ones lie between skip links. As result, the number of calculations required decreases by $k^2$ compared to previous layers. *k* refers to kernel and shows the height and width of the two-dimensional convolutional window.

Balancing all the network dimensions results in a greater efficiency and accuracy. EfficientNet uses a fixed set to improve the performance of CNNs in three dimensions of width, depth, and resolution. Fig. 4 shows a detailed pre-trained EfficientNet network (ver. *B*0). This pre-trained network has 18 convolutional layers in total ($D$ =18), where each layer has a $k(3,3)$ or $k(5,5)$ kernel. Input images have three RGB color channels. This network is trained on 224 × 224 images. The size of the Person Re-identification image set is 160 by 80; the network adjusts EfficientNet to this size to enable the learning.

EfficientNet is more efficient that previous pre-trained models, reducing the computational costs and the demands for a large number of data and accelerating the training process. Fig. 5 shows the application of different versions of EfficientNet to ImageNet dataset in comparison with the application of other pre-trained models.

### B. Transfer Learning

It is necessary to collect enough data to be able to train DNNs. It is hard to collect a large number of data, because the collection of labeled data requires considerable amounts of time and money. Large data usually have a high likelihood of error [18]. Therefore, transfer learning is considered as an efficient way to transfer the knowledge extracted from a source to a target domain [18]; it considerably reduces the network training time and conserves the computational resources. Transfer learning enables us to use the parameters and convolution layer weights of a trained model on a large number of data for our new model on a small number of data. We used the weights of the trained model on the ImageNet dataset, which can be used in Person Re-identification and deliver favorable results as it contains generalized images.

### C. Verification

Two subnetworks extracted the features in the Siamese architecture, and shared the weights. As seen in Fig. 2, The Square Layer can be defined as $f_s = (f_1 - f_2)^2$, where $f_1$, $f_2$ are inputs vector and $f_s$ is the output vector of the Square Layer.

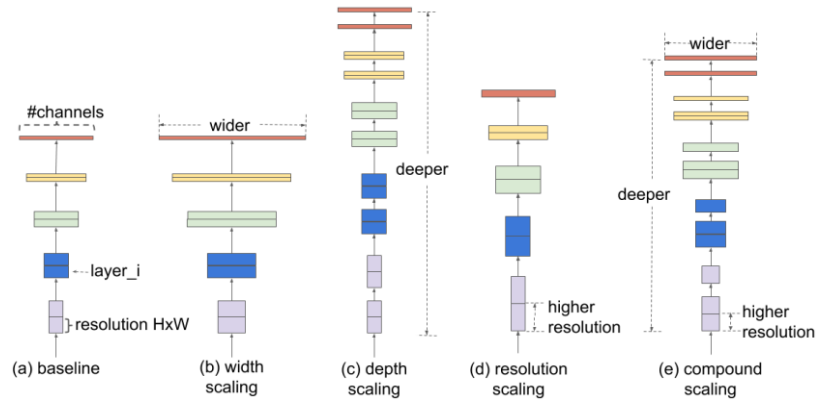

Fig. 3. (a) is an example of a primary network; (b-d) shows the increase in the convolutional scale in width, depth, and resolution in a network dimension. (e) A method of changing the scale of an article that changes all three dimensions with a fixed and uniform ratio [17]

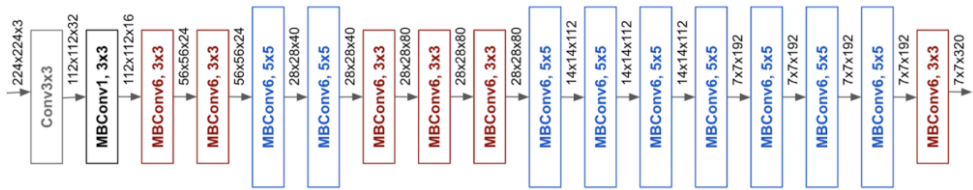

Fig. 4. represents the initial block of EfficientNet B0 [17]

To solve the pedestrian verification problem, we act as a binary classification problem and use the cross-entropy loss function as follows:

$$\hat{q} = softmax(\theta_s \circ f_s)$$
$$verif(f_1, f_2, s, \theta_s) = \sum_{i=1}^{2} -q_i \log(\hat{q}_i)$$

where $f_1$ and $f_2$ are $1 \times 1 \times 4096$, Class $s$ shows the target (similar/different), $\theta_s$ shows the convolutional layer parameters, and $\hat{q}$ refers to the predicted probability. If the input image pair belongs to the same person, $q_1 = 1, q_2 = 0$, otherwise, $q_1 = 0, q_2 = 1$.

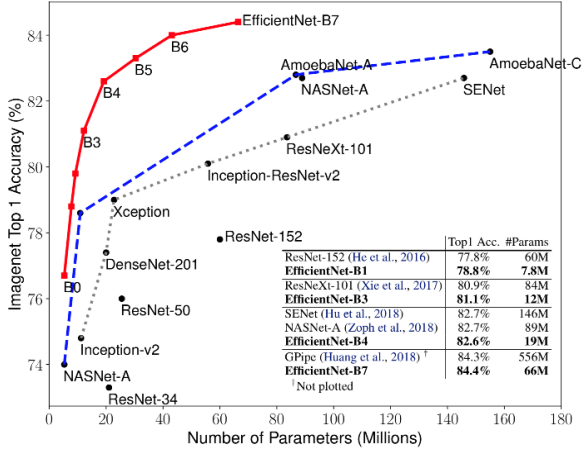

Fig. 5. Comparison of pre-trained models on ImageNet data set [17]

## III. RESULTS

Our proposed architecture was tested on CUHK01 dataset, which is among the most controversial datasets in the area of Person Re-identification. Six versions of EfficientNet, $B$0 through $B$5, were employed to evaluate the performance of the proposed architecture, among which $B$0 showed the best performance. Afterwards, we compared the performance of this model in comparison with the existing models.

### A. Dataset

We ran our tests on CUHK01 dataset, which was recorded and collected at two camera angles. The dataset includes images captured from 971 people, and every person has two images taken by Camera A and two taken by Camera B. Camera A captured the person in full face and Camera B captured them in profile. During training and testing, the image of the person from Camera A was given to the network, and the same image was retrieved from Camera B.

### B. Balancing Positive and Negative Pairs

In each dataset, a person can have four positive image pairs and a large number of negative ones. Overfitting arises due to the smaller number of positive pairs than negative ones. We employed data augmentation to avoid this problem and balance the number of positive and negative pairs. We generated more images using data augmentation techniques such as Horizontal Flip, Zoom, and Width Shifting. As a result, more positive pairs were generated, and balance was created.

The pre-trained EfficientNet model has 8 versions, $B$0 through $B$7. $B$0 through $B$5 were covered because of the hardware limitations. $B$5 is the largest version that can run on our hardware.

### C. Learning Settings

The RMSprop optimization algorithm with the rate of 1e-4 was employed to train the network. Images were given to the network in a 48 batch size and changed to 160 × 80. The gradient descent optimization method was adopted to minimize the cross-entropy error. Training was performed in 18 epochs, and to avoid overfitting, EfficientNet came to use after the last convolutional layer, and the drop-out function was used before the last layer of the network [19]. We ran our tests on Google Colaboratory with Tesla p100 16GB graphics card and 25 GB storage capacity on hard disk drive. Tests were performed on Google Colaboratory and the Keras platform.

This section starts with a comparison of different versions of EfficientNet, and then compares the efficiency of EfficientNet B0 model with that of previously developed methods such as LMNN, KISSME, Quadruplet, Combining, etc. Cumulative Match Characteristic (CMC), shown in Fig. 6, was used to evaluate the methods quantitatively. Table 1 displays the comparison of the proposed methods with state-of-the-art methods on CUHK01.

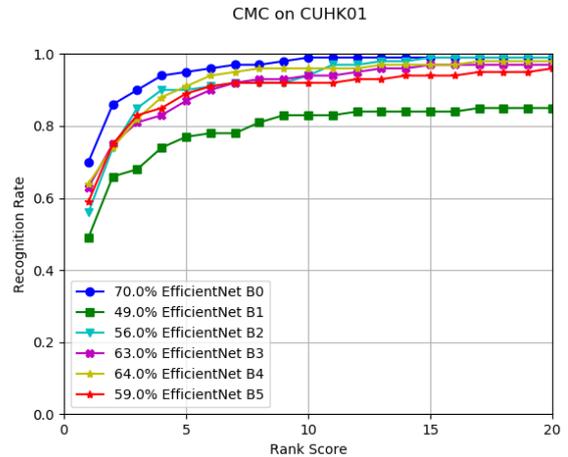

Fig. 6. CMC diagram for the results of different versions of EfficientNet on the CUHK01 data set

Versions $B$0 through $B$5 were used in the proposed architecture. Fig. 6 shows the results from these versions. B0 has the lowest parameter and depth among different versions of EfficientNet. Because of the small number and size of images, the versions with more parameters resulted in overfitting and those with less parameters performed better. In the first order, B0 with 5.3 million parameters reached 70.1%, and $B$5 with 30 million parameters reached 59%. Most of the previous architectures used VGG16 and ResNet50 pre-trained models, having 138 million and 26 million parameters, respectively. The small number of parameters in EfficientNet $B$0 resulted in a shorter training period compared to the reference architectures.

This version also needs a smaller number of images for training, thus reducing the hardware usage.

Table 1. Comparison of the proposed method with the state-of-the-art methods on CUHK01

| Reference | R-1 | R-5 | R-10 | R-15 | R-20 |
|---|---|---|---|---|---|
| LMNN [20] | 13.5 | 31.3 | 42.3 | - | 54.1 |
| KISSME [21] | 29.4 | 57.7 | 72.4 | - | 86.1 |
| DeepReID [2] | 27.9 | 58.2 | 73.5 | - | 86.3 |
| ImprovedDeep [11] | 65 | 89.5 | 93 | - | - |
| LSSCDK [22] | 66.0 | - | 90.0 | 93.3 | 95.0 |
| Atrri [23] | 46.8 | 71.8 | 80.5 | - | - |
| Multi-channel [24] | 53.7 | 84.3 | 91 | - | - |
| E-Metric [25] | 69.38 | - | - | - | - |
| Deep Transfer [26] | 77.00 | - | - | - | - |
| Quadruplet [27] | 62.55 | 83.44 | 89.71 | - | - |
| PN-GAN [28] | 67.65 | 86.64 | - | - | - |
| Combining [29] | 71.5 | 86.5 | 92.5 | - | - |
| Similarity [30] | 68.44 | 86.24 | 93.65 | - | 96.8 |
| Parsing [31] | **83.2** | - | 97.1 | 98.4 | 98.8 |
| Proposed | 70.1 | **95.2** | **99.1** | **99.1** | **99.2** |

## IV. CONCLUSION

Person Re-identification is a problem under the category of image retrieval, which can be solved using deep learning methods. This paper adopted an end-to-end deep learning method to address Person Re-identification, and used an EfficientNet model within the framework of the Siamese architecture to extract the discriminative features. Square Layer was responsible for comparing the features of image pairs. The small number of parameters in EfficientNet accelerates the learning and reduces the need for data, improving the network efficiency and accuracy considerably. This model was experimented on CUHK01 dataset, one of the most controversial datasets in the area of Person Re-identification. Results showed that the proposed method outperforms the state-of-the-art methods regarding the controversial dataset. As a suggestion for future work, the verification and identification model can be used simultaneously to learn the network. Generative adversarial networks can be used to increase the image of any person in the data set. Using the block attention module allows the model to focus on the person in the image and not pay attention to the background.